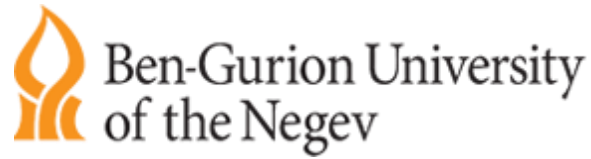

Ben-Gurion University of the Negev
Faculty of Engineering Sciences

Department of Industrial Engineering and Management

# Ph.D. Research Proposal
# Understandable Robots

רובוט "מובן"

**Shikhar Kumar**

Advisor: Prof. Yael Edan

Advisor: Prof. Suna Bensch

Chairman of departmental committee

Date: 18th October 2021

0

**Abstract**

Inspired by recent works in the field of explainable AI the focus of this thesis is on contributing to the development of understandable robots. The goal of this work is to develop a robot equipped with goal-driven explainability, i.e. a robot will explain its behavior to achieve a particular goal in a collaborative setting. The major factor in goal-driven explainability is the human 'theory of mind'. In this work, we will employ Leslies' theory of mind model which includes a mechanical agency, an actionable agency and a belief agency. This thesis will focus on explaining the desire of the robot and the belief of the human if its different to the robot's intention or desire. We aim to develop a common theoretical framework for the development of understandable robots which will include *learning to generate explanations*, *non-verbal and verbal ways of communication* and *explanations in context*.

We premise the explanation will be based on three questions i.e., ***what*** needs to be explained, ***when*** should it be explained and ***why*** an action is being taken by the robot. With these three questions, we will focus on *developing a model for the explanations* which includes clarity, a pattern in which explanations are being communicated to the user, and justification for a particular option. These three questions will provide a basis for *defining different levels of understanding* (corresponding to levels of automation).

*Evaluation will be conducted through a series of user studies performed on different robotic platforms and tasks*. The key performance indicators will include satisfaction, trust, curiosity, fluency, the 'goodness' of the explanation and additional measures which will be assessed through objective and subjective measures. We aim to *develop an aggregated measure to define and evaluate the quality of understanding*.

We designed a preliminary experiment (see Appendix) in which a polite robot was bound to make errors (performing a wrong action while bringing the cubes). Due to the common perception that the behavior of robot is error-free it was observed that the human's perception of the robot was negative when it violated the common belief that the robot would not make the error. This experiment will be further extended to include different levels of understanding and explanations will be generated to mitigate the error.

A second experiment will be designed in which a robot learns to generate the explanations. In this experiment we will investigate the ***what*** and ***how*** much information needs to be uttered for a collaborative task. A collaborative task will be designed in which a robot will plan to arrange a set of cubes according to geometrical shapes and would generate explanations in a syntactic manner. The disparity between the mental model of robot and the estimated model of the human will be treated as the state for the learning algorithm and the human feedback will be the reward function. The explanation will include the contextual as well as the historical aspect. In this study, investigation will focus on the comparison of utterances which involve spatial relations, object properties and object placement with respect to the humans frame of reference.

For making the interaction more natural, inclusion of non-verbal aspect of interaction will be investigated. From the previous experiment the robot will focus on learning the different aspects of the non-verbal communication such as hand and gaze movements. The robot will learn how much utterances backed by the non-verbal communication should be communicated. Here a user study will be conducted with a between the subject design in which we will compare between verbal communication and non-verbal communication design.

Finally, the work will include an investigation of the contextual form of explanations. In this study, we will include a time-bounded scenario in which the different levels of understanding will be tested to enable us to evaluate suitable and comprehensible explanations. For this we have proposed different levels of understanding (LOU). A user study will be designed to compare different LOU for different contexts of interaction. A user study simultating a hospital environment will be investigated.

We expect this research to contribute to the novel field of developing understandable robots while contributing with specific design guidelines and  metrics for evaluating the quality of understanding.

# Table of Contents

# 1 INTRODUCTION

As robots become more and more capable and autonomous, the use of robots in daily tasks by nonprofessional users and bystanders will increase. For smooth and efficient interaction, robots have to be designed such that their behavior and states are understood by the interacting humans.

Understandability and explainable behavior have been defined in various XAI literature (Barredo Arrieta et al. 2020). *Understandability* is defined as the notion of communicating the agent's function (how it works) which could help a user interacting with the agent to comprehend its function. Understandability refers to explaining the agent's functionality without revealing any technical details. *Explainability* is defined as the interface between human and robot which helps in making the decision of agents comprehensible to the user. This proposal is part of a larger project which focuses on aspects related to *What* (the amount of information that needs to be conveyed for robotic action) *When* (the moment at which the communicative action needs to be generated), *How* (the medium to dissipation of the communicative actions) and *Why* (the justification for a particular action of the robot) to increase the understandability of robots working in collaboration with humans via efficient communication during interaction.

In (Hellström and Bensch 2018) a model for understandability is proposed. It describes the state of mind (SOM) of a human ($M_H$) and a robot ($M_R$). $M_R$ contains a model $m_H$ of $M_H$, and $M_H$ is assumed to contain a model $m_R$ of $M_R$. *Communicative actions* by the robot aim at reducing the disparity $|M_R - m_R|$ between the robot's mind and the human's model of the robot's mind. For example, a robot may flash a red light to signal to a bystander that it stops because of an obstacle on floor. In this example the robot senses the obstacle which is part of $M_R$ and the user has a model that the robot would continue moving ($m_R$). Hence, there is a disparity between $|M_R - m_R|$ and hence a communicative action (flashing a red light) has been generated.

Some robots make decisions based on opaque black box models, which may lead to distrust among users due to lack of understanding of the robot's actions. Techniques for *Explainable Artificial Intelligence* (XAI) aim to provide explanations of the black box models (Doran et al. 2018)(Chandrasekaran et al. 2017). Such explanations typically describe how certain input data contribute to a specific decision. Inspired by XAI many researchers have explored the idea of *goal driven explainable intelligence* (XGDAI), in which the focus is on explaining the robot's action. XGDAI exploits the idea of **Theory of the Mind** (ToM) of the robot, and to comprehend the robot's behavior, which in turn leads to improved human-robot interaction which can lead to improved coordination between the human and robot.

## 2 THE RESEARCH

### 2.1 Research Objective

The objective of this work is to develop state-of-the art methods for increasing the understandability of robots in collaborative tasks with humans. The research will incorporate the theory of mind model (based on Leslie's theory of mind model; Baron-Cohen et al. 1985, Leslie 1987, Leslie 2010) of belief, desire and action or goal. Following are different aspects of understandability which this thesis focuses on:

1) Developing a learning algorithm to generate verbal explanations of the plans made by the robot to humans. This would help the robot to adopt the best possible explanation for its plan. The robot would in turn learn about the state of mind of different users and would adapt its explanations depending on the different users.
2) Developing combined verbal and non-verbal communication for better understandability. The inclusion of non-verbal communication such as gaze movement, eye movement, pointing to objects would make the interaction more natural and efficient.
3) The inclusion of context in an explanation i.e., different explanations required for different types of tasks and environments.

### 2.2 Research Significance

Automation in society has greatly improved the interaction of humans with different robots. With the advent of automated robots in the society, there is a need to develop a robot which explains its plans and acts (Fong et al. 2003). There would negative effect on the interaction quality between the user and robot if the problem is not addressed (Bensch et al. 2017). Anxiety is developed in HRI similar to human-human interaction when robot is unable to explain its action (Nomura, 2006). In collaborative tasks, the performance and interactivity between the collaborators decreases if the robot is unable to communicate its action (Gielniak and Thomaz 2011) (Cingolani 2014). In a non-collaborative task, the safety of the human being is a big concern if the robot is unable to explain its action and would facilitate in collision between the robot and the bystander human (Hellström and Bensch 2018). The user would lose its trust when it cannot understand the behavior or decision (Miller 2019). Various researchers agree that social agents should provide some degree of interpretability so that it could well be comprehended and to enhance the trust levels (Lomas et al. 2012; Graaf and Malle 2017).

It is important that the users understand what the robot is doing. This requires both verbal and non-verbal cues to help the user understand the robot's plans. The critical component of understandability is the nature of messages that must be communicated such that humans are not burdened by huge cognitive load. In a successful collaborative task between a human and a robot, there needs to be transparency as well as the need to communicate precise and clear enough information for each agent to understand each other plans. Moreover, following are some of the major critical issues in real world robot understanding domain which must be addressed:

1) The adaptation of different methods or types of verbal communication when the user is unable to understand the robot.
2) The capability of the robot to predict the mental model of the user for increased understanding
3) The development of non-verbal cues required for understandability of the robot.

### 2.3 Expected Contribution and Innovations

There have been numerous researches on XDGAI detailed in the survey report (Sado et al. 2020) focusing on deliberative (navigate through behavioral spaces, act like they think, and predict the consequences of an action), reactive (reacts to the changes in the environment) and hybrid (combination of deliberative and reactive agents' techniques) architectures. However, there are numerous elements lacking in the current research domain. Few factors which can improve understandability are natural language processing for the generation of explanations (except in (Singh et al. 2021), introduction of ToM for agents to understand the perception, precise explanation and integrating verbal and non-verbal mode of communication such as gaze movement and gestures. Further, there is need to develop understandable robots based on context of the environments. For example, fast explanations are required in a collaborative task for rescue operations, accurate explanation is more important in the case of a medical surgery, in other cases perhaps a long and very detailed explanation could be transmitted if time is not a constraint, aiming to increase the humans understanding for example in a social robot interacting with an older adult.

## 3 THEORETICAL BACKGROUND AND LITERATURE REVIEW

### 3.1 Theory of mind model

Leslie's theory (Baron-Cohen et al. 1985) (Leslie 1987) (Leslie 2010) assumes that the central processor of information in infants evolves as an architecture reflective about the world's properties. Upon this assumption the knowledge about the world's properties could be classified in three main agencies namely mechanical, actional and attitudinal. Leslie further suggested these agencies are not hierarchical in nature with respect to development of sense in the infants rather different agencies are the subsystem and could develop in parallel. The modules that deal with the mechanical agency are termed as a theory of body module which can be explained by the rules of mechanics. Theory of body is responsible for early development about the knowledge of physical structures in infants. Its goal is gathering the knowledge about the mechanical description about an object and the motion they enter into. We do not consider the theory of body in this work as in a collaborative task environment the robot needs to explain about its plan of action to achieve a goal rather than explaining robot's internal working and is beyond the definition of understandable robot. The module that deals with actionable and attitudinal agencies comes under the cognitive structure of theory of mind. The theory of mind involves in understanding the intentional properties of the agent. The intentional properties might include the desire to achieve a goal and react to the distant environment (perception). The agency which explains the events based on the goals or the desire of the agents i.e., actions of the agents is termed as actionable agency and Leslie termed it as System-1. This may include either requesting from the other agent to achieve a goal or comply to the orders of the other agent's goal. For example, in a table setting task with humans and robot, when a human instructs the robot to bring a glass, the human believes that the robot would approach the cabinet full of glass and bring the glass for setting up the table. The attitudinal agency referred to as System-2 of theory of mind explains about the development of the meta-representation (M-representation) i.e., the ability to pretend and understand the pretense in others. It is helpful in understanding the belief held by a person different from either our own knowledge or knowledge obtained from observable world. In the same example, when a human instructs the robot to bring a glass, the robot slowly erroneously approaches the cabinet full of plates while the human is in belief that robot would bring the glass for table setting.

### 3.2 Related work in understandable robots

In the field of understandable robots, a theoretical model was proposed by (Hellström and Bensch 2018) based on the requirement of generating communicative actions when there is disparity between the robot's mind and human model inside the robot's mind. The communicative action was based on *what* information needs to be communicated, *why* an action or plan has been decided and *how* the robot should communicate its explanation. The challenges of generating explanations should consider the fundamental element of sense making (Papagni and Koeszegi 2020). The authors further suggested that explanation should include iterative type of communication, contextual form of explanation generation and combination of using non-verbal and verbal cues. Both studies lacked user studies and were limited to presenting a theoretical framework for understandable robots.

A previous review work (Sado et al. 2020) in the field of GDXAI categorized according to the behavioral aspects of interaction between the agent and the human i.e., deliberative (where agents plans ahead to achieve goal), reactive (agents respond to the environmental changes) and hybrid model (combination of reactive and deliberative actions). The goal driven action plan generates explanations when an agent finds a mismatch between the expectation of a plan and the current status by tracking the agents behavior (Nau 2007; Molineaux et al. 2010; Jaidee et al. 2011). The belief, desire and intention (BDI) model explains based on underlying belief and desire (Georgeff et al. 1999; Malle 1999; Harbers et al. 2010; Van Camp 2014). The main aspects is to explain the human errors (Malle 1999). The situation awareness model is built on the BDI model which had built an interface communicating information not only about current status and reasoning but also on future projection (Boyce et al. 2015; Chen et al. 2018). In a proactive explanation model the agent explains the surprise element of its action proactively such that participants are not flabbergasted (Gervasio et al. 2018). These are highly domain specific and have limiting application.

The automated rationale generation model trains the encoder-decoder neural network to generate the explanation of behavior of the agent as if a person explains to another (Sequeira and Gervasio 2019). The explainable reinforcement learning way to generate an explanation allows them to learn the policy to explain the behavior based on trial and error (Sequeira and Gervasio 2019). Generation of explanation for robotics failures has been addressed by invoking an explainable AI model such as action-based, context based and history-based explanation (Das et al. 2021). To explain robot action a binary tree was used to generate explanations (Han et al. 2021a). The progressive explanation generation algorithm has proved to increase the performance of the task in a scavenger hunt and escape room task (Zakershahrak et al. 2021). This study has considered the mental model of the human being as the state of reinforcement learning and a reward function was generated by the inverse reinforcement learning retrieved from the human's preference.

Another study which focused on comparing non-verbal and verbal communication suggested that the non-verbal mode alone is not sufficient for explaining the robot's actions or plans (Han et al. 2021b).

## 4 A CONCEPTUAL MODEL FOR UNDERSTANDABILITY

### 4.1 Overview

The framework for understandability we propose in this thesis includes three main aspects: theory of mind model (Section 4.1), different parameters influencing the understandability (Section 4.2) and development of the level of understanding (section 4.3). Based on these aspects of understanding we have formulated the research questions which will be explored in the study.

### 4.2 Disparity in Mental Model

The state of mind of an agent could be loosely defined as it's action with respect to goal, desire, beliefs, intention, emotions, capabilities and task progress and uncertainty. To develop an understandable robot model, the robot needs to communicate when there is a disparity between the state of mind of the robot ($M_R$) and estimated model of robot's mind ($m_R$) by the human. The robot state of mind ($M_R$) would consist of components such as percepts of environment, belief, desire, intention, goal, limitation and capabilities. The percepts of environment could be developed with existing state-of-the-art technology machine learning algorithm. The theory of mind ($m_R$) will consist of components such as interacting environment including humans and it's actionable and attitudinal agency. The disparity between |$M_R$ – $m_R$| could be described as perception of robot towards the environment and model of human state of mind. In this work, we propose two disparities with regards to actionable agency and attitudinal agency. Fig 1 depicts the two different levels of disparity in attitudinal and actionable agency.

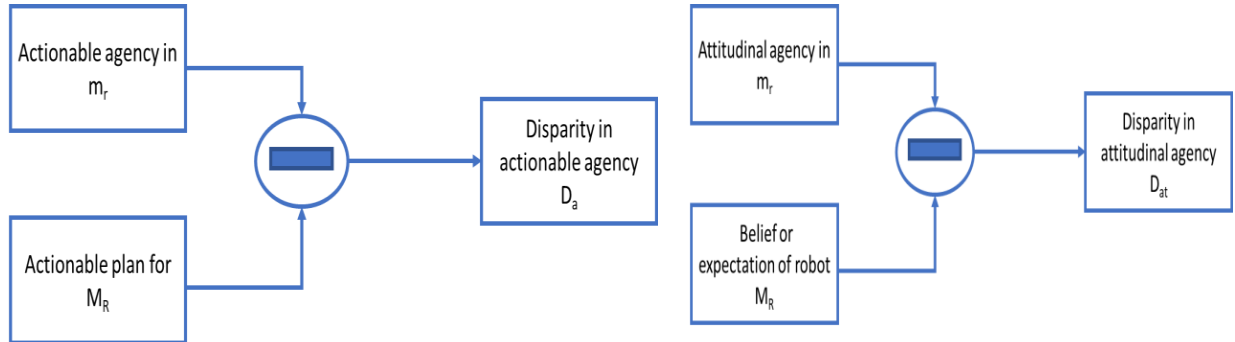

Figure 1. The two different disparities between the robot and interacting environment in the actionable or attitudinal agency. The arrow in this figure shows the interaction between the models that generate a disparity.

## 4.3    Parameters for Understandability

The following parameters should be considered for development for generating explanations for the understandable robot (Fig. 2):

1. a context of the explanation,
2. a learning method to generate the explanation, and
3. verbal as well as non-verbal ways of communication.

This work will focus on all these three parameters aiming to develop an understandable robot with the ToM model demonstrated in the previous section. Each of these developments are detailed below.

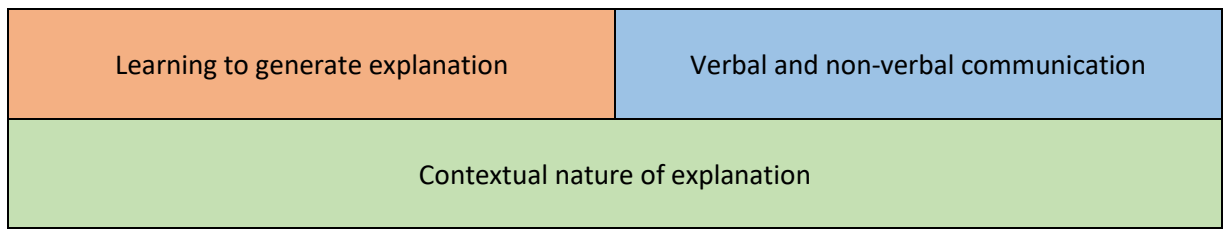

Figure 2. Parameters for explanation generation

### 4.3.1   Learning method to generate explanation

The explanation should be precise and clear enough to able to communicate the intention of the robot. However, it is difficult to have a general explanation for all collaborative tasks and for all users. There is a need to adjust the explanation in a different manner such that it satisfies the specific user and the explanation is easily comprehensible. This work will focus on the development of a syntactic way of generating natural language. To achieve this, we will develop a learning algorithm which will allow the agent to learn the appropriate explanations which need to be communicated. The disparity between $D_a$ or $D_{at}$ or both would be states of the learning algorithm. The reward function will be the feedback of the human i.e., if the user does a mistake this will yield a negative reward and the state would change to next state and it would generate an explanation with a higher granularity.

### 4.3.2 Verbal and non-verbal communication

In human robot interaction inclusion of verbal commands along with gestures or gaze movements could make the interaction more natural and expressive. In this work we will focus on two levels of communication i.e., one will be with only verbal communication and the other will be a combination of non-verbal communication and verbal communication. In the non-verbal communication we will focus on movements such as eye movements, gaze movement and light blinking, and pointing to the object with pen etc.

### 4.3.3 Context of explanation

The explanation of the robot will also depend on the context of the environment in which the robot navigates of the robot. For example, learning the way to generate explanation would not be feasible during a task which requires urgency. This work will focus on developing a common framework for different levels of understandability These different levels of understanding will help the robot to choose the explanation according to the situation. A theoretical framework of different levels of understanding (Section 4.4) will communicate information based on the what, why and when information needs to be communicated such that the explanation of the robot's plan is easily comprehended and it is coherent with the belief of the human.

## 4.4 Different Levels of Understanding

We propose different levels of understanding (Table 1) the robot must adopt while communicating with humans based on the three questions i.e., *What* information about the desire or action or plan should be communicated to the human? *Why* such desire or action or plan have been selected by the robot? *When* should robot be communicating these plans? Based on these three questions we have modeled three different aspects or understanding i.e., clarity, justification and explanation patterns. These three modalities are detailed below.

### 4.4.1 Clarity

The clarity in explanation could be defined as explaining the action or intention of the robot and changing the contrary belief of the human. In this work clarity will be defined in two levels i.e., low level of clarity and high level of clarity which are as follows:

Low level of clarity: In the low level of clarity we would only consider explaining the what the desire is? The robot would explain the plan to the user irrespective considering the belief of the user.

High level of clarity: In the high level of clarity the robot would also consider the belief of the human and clarifying the belief of the human if it is contrary to robot's intention as well as explaining the intention of the human being.

### 4.4.2 Justification

The robot would have to justify its desire or action to the user. The plan to reach goals for the robot and human could differ, therefore the robot needs to justify its plan to the user. The justification of its action can be divided in two levels i.e.,

Low Level of justification – The robot would not justify the user its course of action or plan.

High Level of justification – The robot would justify the user its course of action or plan.

### 4.4.3 Explanation pattern

Ideally explanation should be generated either if the human is not aware of the intention of the robot or the belief of the human about the task is not in coherence with the intention of the robot. This is divided into two levels:

Static pattern – The robot would explain the user every time when it is about to act.

Dynamic patter – The robot would generate the communicative action as per the requirement of the user.

The different aspects of understanding as described above have been combined to develop different levels of understanding (LOU). The Table 1 describes the different levels of understanding. The low LOU is when all the three-aspect described above is low (represented as 0 in Table 1) and the highest LOU is when all the three aspect would be high (represented as 1 in Table 1).

## 4.5 Research questions

With the model enumerated above, this work will focus on the following research questions:

a) Should the explanation be generated when the belief of another agent is contrary to the agents' action?
b) Which different levels of explanation should be included to change the belief of the other agent?
c) Is an iterative way of generating explanations more comprehensive compared to a single generation of explanation?
d) Does combining non-verbal communication with verbal communication improve understandability as compared to only verbal communication?
e) How should a robot adapt to the different levels of understanding depending upon the environment/task/user?

**Table 1. Different levels of understanding**

| Levels | When | Why | What | Description | Explanation |
|---|---|---|---|---|---|
| Level 1 LOW LOU | 0 | 0 | 0 | Intention explanation in a static manner | The robot would only inform about the different plans from which user can choose |
| Level 2 | 0 | 0 | 1 | Intention explanation and eliminating false belief in a static manner | The robot would choose the plan and will tell the user about it |
| Level 3 | 0 | 1 | 0 | Justification of Intention explanation in a static manner | The robot would try to give justification for each plan generated for each task |
| Level 4 | 0 | 1 | 1 | Justification of Intention explanation and false belief in a static manner | The robot would justify the plan generated by it. |
| Level 5 | 1 | 0 | 0 | Dynamic pattern of decimating intention explanation | The robot would tell about different plans to user when it demands |
| Level 6 | 1 | 0 | 1 | Dynamic pattern of decimating intention explanation and removing false belief | The robot would tell about plans selected by it to user when it demands |
| Level 7 | 1 | 1 | 0 | Dynamic pattern of decimating intention explanation with justification | The robot would enumerate all the plans available and justify each one of it when the user demands |
| Level 8 HIGH LOU | 1 | 1 | 1 | Dynamic pattern of decimating intention explanation and eliminating false belief with justification | The robot would justify the plan selected by it robot and would justify its actions |

# 5 METHODOLOGY

## 5.1 Overview

Based on the initial framework of theory of mind and the conceptual framework of understandability we will test the importance of different parameters required for generating explanations. We will try to find out the importance of each parameter for understandability for developing an understandable robot. For this we will use different robotic platforms (manipulators and mobile robots) equipped with simple natural language processing capabilities in a series of user studies in different robotic tasks as detailed below. The analysis and results from these user studies will help us develop a prototype for an understandable robot based on the different contextual environments. A sequence of experiments is planned (Figure 3, Table 2) in which each experiment tests a subset of influencing parameters which feeds into the next experiment.

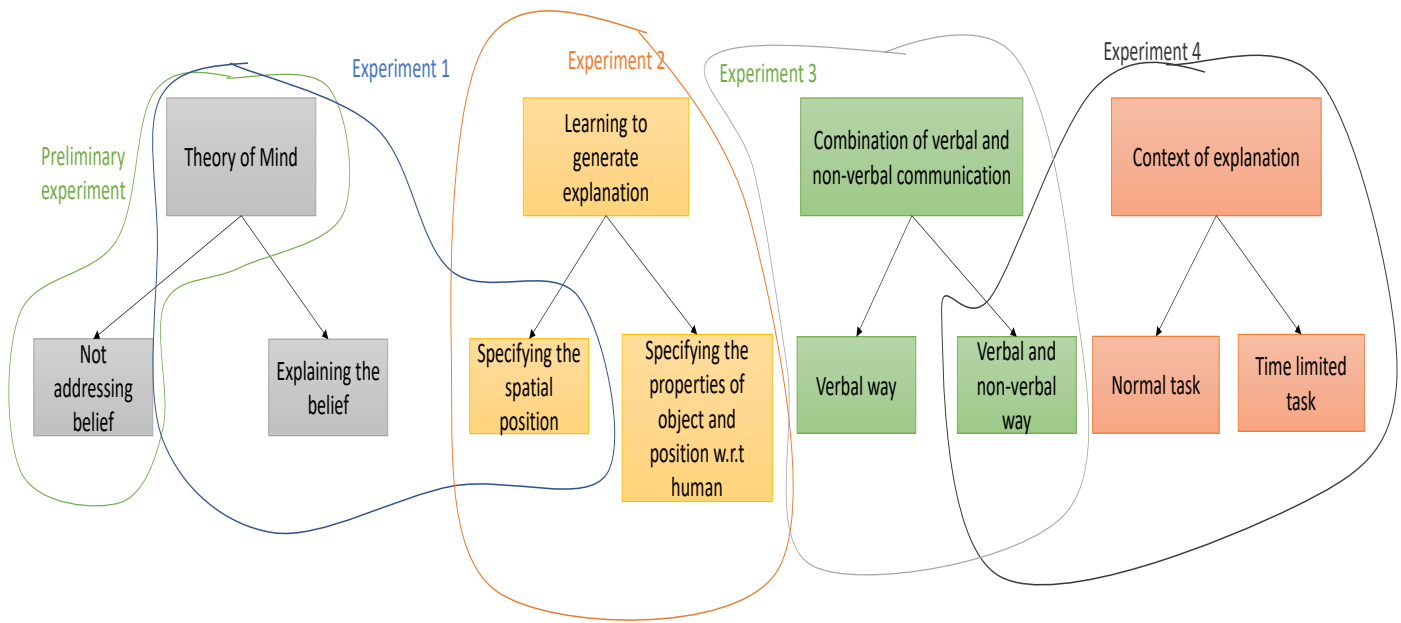

Figure 3. Planned future experiments with different parameters of understandability.

**Table 2 Main Goals of Each Experiment**

|  | *Main Goal* |
|---|---|
| *Experiment 1* | Development of non-iterative way of explanation for the errors made by the polite and non-polite robot. |
|  | Comparison between No explanation given for the error made to explaining the cause of error to the user |
| *Experiment 2* | Investigate *what* information and *how much* information a robot has to verbalize in a given context in order to convey its intention or plan to the human. |
| *Experiment 3* | Development of way of communication including the non-verbal's aspect of communication which includes gaze, eye movement, finger pointing etc. Comparison between the different non-verbal communication modes (which would include verbal communication) with the only verbal communication. |
| *Experiment 4* | Development of time limited model of understanding using the different levels of understanding |
|  | Comparison of different levels of polite behavior should be employed in the context of time limited interaction |

## 5.2 Background Questionnaires

The preliminary background check in all user studies will involve technology assessment and propensity questionnaire (TAP) (Ratchford and Barnhart 2012) and negative attitudes towards the robot scale (NARS) (Nomura et al. 2006) to judge the background knowledge of participants about the new technology and attitude towards the robot.

## 5.3 Key Performance Indicators

1) **Fluency of interaction:** Fluency in human-robot collaboration helps in encouraging the teammates and helps in building great coordination between the human and robot (Hoffman 2019). The objective measure for fluency in understandable robot could be robot idle time, human idle time and task duration between the concurrent activity and functional delays. Human idle time defined as the percentage of time when human was not active. Robot idle time is defined as percentage of time robot was not active. Concurrent time would be defined as percentage of time both the agents were active in the interaction. Functional delay is time delay between completion of task by one agent and beginning of the task by another agent. The subjective measure could be taken from subset of questionnaire from (Hoffman 2019). The main hypothesis is that a more understandable robot would be more fluent.

2) **Adequacy of explanation:** The property of good explanation depends upon the user needs and the knowledge the user gains from the explanation. The objective measure could the smaller number of times the user would urge the robot for explanation. For iterative explanations, this could be considered as the level at which a user would ask for explanation. The subjective measures will be taken from the parts of questionnaire from (Hoffman et al. 2018).

3) **Explanation satisfaction:** Explanation satisfaction would be defined as the user needs to get all the appropriate and satisfactory explanation to understand the robots action (Hoffman et al. 2018). Objectively it could be judged if the number of time user ask the experimenter about the explanation and also time required to respond to robot's explanation. The subjective measures will be taken as subset from the questionnaire (Hoffman et al. 2018).
4) **Measuring Curiosity:** Measurement of curiosity helps in evaluating the eagerness of human to know more about the explanation of the robotic task (Hoffman et al. 2018). The subjective measure could be defined from the questionnaire (Hoffman et al. 2018).
5) **Trust:** Trust could be defined as the degree to which the human has reliance and satisfaction on the robot. Trust in understandable robot would define as the degree to which a user can rely on the explanation given by the robot (Hoffman et al. 2018). The objective measure could measure from perception of safety i.e., proximity to the user. The subjective measure will be taken from the questionnaire (Hoffman et al. 2018).
6) **Aggregate measure:** Both the objective and subjective measure (Table 3) could be normalized from which then we calculate the average of these measures (Olatunji et al. 2021).

    N.M. = $(V_i - V_{min})/(V_{max} - V_{min})$

    Where $V_i$ is the value of measures and $V_{min}$ and $V_{max}$ is the minimum and maximum value of the metric and N.M. is the normalized value.

    A.M. = $(1/N) \Sigma$ N.M.$_i$

    Where A.M. is the aggregate measure and N is total number of each measures.

**Table 3 Questionnaire required for evaluating different quality of understanding parameters**

| Measure | Questionnaires |
| --- | --- |
| **Fluency of interaction** | 1) I felt information provided by the robot was at right time <br> 2) I felt action of the robot was at right time |
| **Goodness of explanation** | 3) The explanation helps me understand how the robot works <br> 4) The explanation of the robot sufficiently detailed <br> 5) The explanation is actionable, that is, it helps me know how to use the robot |
| **Explanation satisfaction** | 6) This explanation of how the robot works is satisfying. <br> 7) This explanation of how the robot works is useful to my goals. |
| **Measuring Curiosity** | 8) I actively seek as much information as I can in a new situation <br> 9) I feel stressed or worried about something I do not know |
| **Trust** | 10) I feel safe that when I rely on the robot I will get the right answers. <br> 11) I am confident in the robot. I feel that it works well. |

5.4 **Participants**

For each experiment we will try to recruit at least 45 students with engineering background and 45 students from non-engineering background. The diverse population is important so we include feedback from non-technical users. This will help us to ensure the understandable robots developed are fit for both the expert and non-experts. In each of the experiments different participants will be recruited to avoid bias from previous experiments.

## 5.5 Analysis

All the key performance indicators measures will be defined as dependent variables. The different parameters related to understanding, level of understanding, gender of participants, background of participants and age (if we conduct experiments with different aged populations) will serve as independent variables. The data obtained from the experiments will first be checked for its statistically normality. If the data is normal, then it will be analyzed using general linear mixed model regression otherwise we will employ a cumulative linked mixed model.

## 6 DETAILED EXPERIMENTS

### 6.1 Preliminary Experiment: User perception of polite robot which can make mistake

A preliminary study (detailed in Appendix A) focused on whether there is a need for explanation in collaborative robot task when the belief of the human is different about the action of the robot. The media portrays robot to impeccable (Bruckenberger et al. 2013). This influences and helps in building a mental model of a flawless in robots especially in the minds of people who have never interacted with robots (Mirnig et al. 2017) . In this work we focus on developing a polite robot which can err. To evaluate this, we developed a polite robot based on the socio linguistic theory of polite behavior (Lakoff polite rules). We then developed four levels of interaction (detailed in Appendix A) and evaluated the following research questions:

1 Do robot's need to generate explanation to human collaborator if robot makes error in action?

2. What is the effect of user's perception with polite and non-polite robot making error in action?

Design: A human-robot collaborative tasks was designed (Appendix A) with polite and no polite behaviors which makes a mistake in action one of the scenarios. Thirty participants experimented with the system in a within the participant experimental design.

The Friedman test revealed significant differences for all the dependent variables i.e., enjoyment ($\chi^2$ (3) = 38.37, p <0.001), satisfaction ($\chi^2$ (3) = 48.60, p <0.001) and trust ($\chi^2$ (3) = 46.46, p <0.001). Detailed results are presented in Appendix A. The results of this preliminary experiment suggest that the user's perception of a robot that makes errors in action, irrespective of politeness, is negative. Our hypothesis is that this can be improved by acknowledging the error and making the robot explain its behavior when it makes mistake. Future work (described in experiment 1) will focus on evaluating different levels of understanding for resolving the errors in a similar experimental design.

### 6.2 Experiment 1

Goal: Development of explanation for the errors made by the polite and non-polite robot. Comparison between two robots – one with 'No explanation' given for the error; the second with a robot that explains the cause of error to the user.

Design: We will evaluate different parameters of the understandable robot to study their influence. The framework used in the preliminary experiment would be extended with addition to the explanation with different levels of understanding. In this experiment explanation would be added when the error is made by the robot. The explanation will include the context or the history of the error making and how to overcome the error in a collaborative task. The type of error here in this work is wrong action performed by the robot. To generate the explanation, we could use the test of two different LOUs i.e. LOU level 2 and level 4. In the preliminary experiment there was no explanation to the errors. The explanation will also include the non-verbal communication such as showing the error making robot as blinking of red light on the robot or the screen to depict error. The experiment will be a between the subject design in which each of the participant would get the different explanation for polite robot making errors.

Research questions:

1) Would inclusion of the explanation which only has clarity help in improving the perception about the error making robot?
2) Would addition of non-verbal component of explanation help in quality of explanation?
3) Would inclusion of justification with clarity about the error help in improving the perception about the error making robot?

### 6.3 Experiment 2

Goal: Investigate *what* information and *how much* information a robot has to verbalize in a given context in order to convey its intention or plan to the human.

Design: We will consider a collaborate task where the robot instructs the human to arrange different shapes of object on the table. The user and robot are sitting across the table where the boxes are kept. The robot would explain the user using syntactic template which may consist of context of the action (such as the location of the object respective to human or the robot or properties of the robot) and the history of the utterances (such as if the previous utterance is to move the circle to the left then the next utterance could move it forward). A user study will be conducted to investigate the human reaction to the comparison of utterance of the spatial position of the object (like left, right, next to etc.) versus the properties of object (such as color of object, shape of the object) and when the robot explains human perspective.

The robot first generates a sequential and incremental plan for arranging the boxes, and then instruct the human to move them, one by one. The robot has syntactic templates to generate natural language utterances of varying complexity. Suppose that the circle has to be moved from its place as shown in figure 4. The following utterances by the robot are possible:

Step 1:

1."The circle"

2."Move the circle"

3."Move the circle on your left"

4."Move the circle on your left to below the square in the middle row"

After reaching the state 1 in Figure 4, the robot could utter following remark to reach state 2.

Step 2:

5. "Exchange the position of the circle with triangle"

6. "Exchange the position of the circle with triangle in the last row"

7. "Exchange the position of the circle with triangle in the last row with respect to you"

With the above possible utterances and further similar utterances the robot will try to collaborate with human to arrange the geometric shapes.

If the human makes a mistake or does not understand after first instruction is given, the robot would utter a verbal instruction with higher complexity.

The robot not only correlates the human's reaction to what it said but also learns how much to say so that the human understands what the robot expects. The learning algorithm will not only correlate actions and observations but also learn causal relations (e.g. when the robot says x, the human reacts with y). The learning algorithm also allows the robot to create a mental model of the human, estimate the discrepancy between the mental model of the human and the robot's expectations and learns which communicative actions (i.e. utterances) to say to reduce the mismatch between robot and human.

Research question:

1) How much and what information should be uttered so that human can understand the system?

### 6.4 Experiment 3

Goal: Development of learning algorithm for communicating including the non-verbal's aspect of communication which includes gaze, eye movement, finger pointing etc. Comparison between among the different non-verbal communication (which would include verbal communication) with the only verbal communication.

Design: In this experiment we will include and compare between different types of combinations of non-verbal and verbal iterative method and the only verbal iterative method. In previous example the robot would shake its head in a negative way in combination with shut eye when the user makes a mistake and in one point it would point at the object with its hand. For example

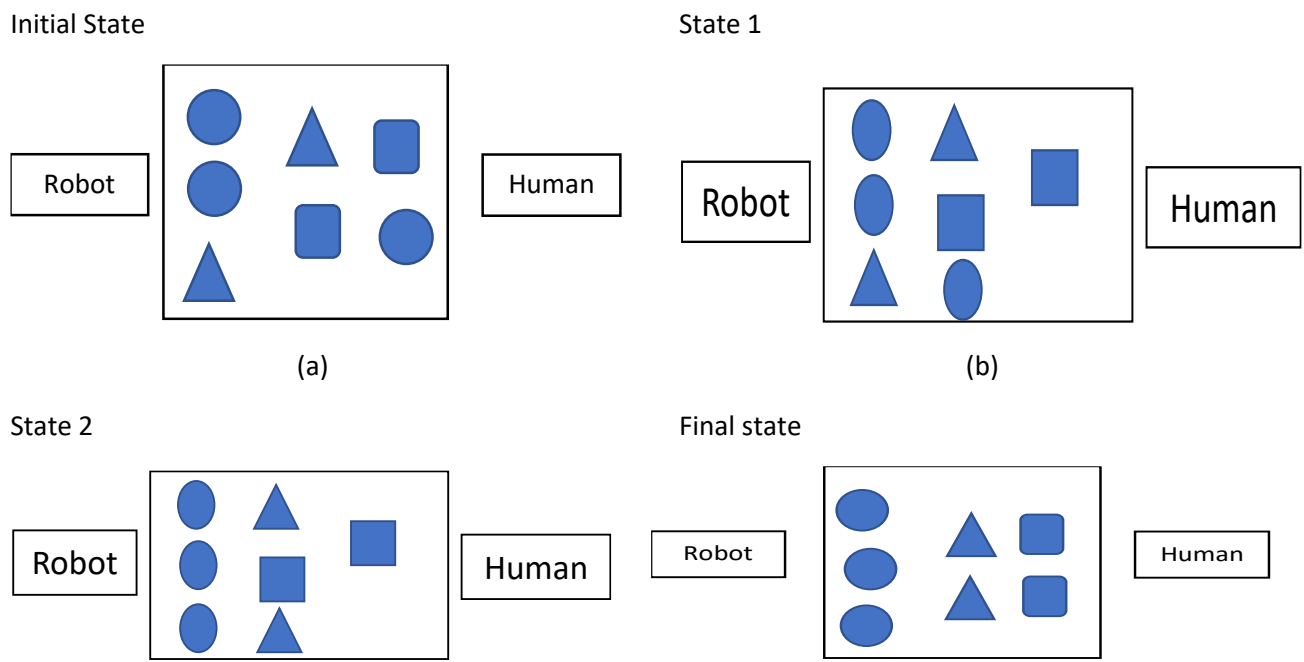

Figure 4. A possible scenario in this experiment

Move this circle (Robot would point to the particular circle with its hand)

## 6.5 Experiment 4

Goal: Development of time limited model of understanding using the different levels of understanding. Comparison of different levels of polite behavior should be employed in the context of time limited interaction

Design: In this experiment we will take into consideration the time factor when generating an explanation. This will include a teleoperated mobile robot which has to go to various patients from two scenarios i.e., one is from a normal ward and another is from Intensive care unit (I.C.U.) As we know in the normal ward patients are less critical so the time limits of the task is not a factor so the robot will explain the user to navigate through the hospital setting and avoid obstacles. However, in the I.C.U regular check-up is required with limited time to explain the operator if the situation is wrong. So, we will try to emulate the environment and check the different levels of polite behavior required to teleoperate the robot in the hospital.

**Appendix A**

# Effects of collaborative robot's politeness and errors on users' perceptions

Shikhar Kumar [1*†], Eliran Itzhak[1†], Yael Edan[1], Galit Nimrod[2], Vardit Sarne-Fleischmann[1] and Noam Tractinsky[3]

*Abstract—* **We incorporated polite skills of human-human interaction into a collaborative robot and investigated the influence of politeness and of erroneous robot behavior on users' perceptions of the robot. Polite behavior for the robot was designed according to Lakoff's theory of politeness. Two robot conditions were designed, in one of which the robot behaved according to politeness rules while in the other the robot's behavior was strict (yet not rude). Erroneous robot behavior was manipulated by designing errors into one condition while the other condition remained error-free. Results revealed that polite behavior had no significant impact on how users perceived the robot. However, wrong actions performed by the robot had a negative impact on the interaction. 68.32 % of the respondents rated the erroneous polite robot as the least preferable. However, the polite behavior with a robot performing correctly was preferred by 52.36% of the respondents.**

INTRODUCTION

With the increasing usage of robots in diverse fields, socially skillful robots equipped with high-quality human-robot interaction (HRI) are important to ensure efficient collaboration (Loi et al. 2018)(García-Soler et al. 2018). An important social aspect of human-human interaction is politeness.

Politeness helps increase cooperation and avoid conflicts between individuals. Reeves and Nass's Media Equation theory (Reeves and Nass 1996) proposes that people respond to computers as they respond to other human beings. Nass (Nass 2004) further empirically supports the argument that people tend to behave politely towards the computer after the script of conversations has been initiated. However, they did not investigate the aspect of people's expectations of polite behavior. Other studies (Whitworth and Liu 2009)(Whitworth 2005)(Hayes and Miller 2010) suggested designing an etiquette interaction between humans and computers. However, their theoretical analyses were not examined empirically.

A theoretical framework for defining politeness in the field of human-computer interaction (HCI) has been developed (Bar-or and Tractinsky 2018), inspired by Grice's seminal work (Grice 1978) on the rules of conversation and Lakoff's theory (Lakoff 1973) of polite behavior. Bar-Or and Tractinsky (Bar-or and Tractinsky 2018) have also demonstrated empirically that polite behavior had a positive effect on efficiency and user perception. The current study was inspired by this line of research. Lakoff (Lakoff 1973) has defined politeness "*a system of interpersonal relations designed to facilitate interaction by minimizing the potential for conflict and confrontation inherent in all human interchange.*" Lakoff's polite theory goes beyond Grice's work to include three sub-rules, namely 1) Don't impose, 2) Give options, and 3) Be friendly. Lakoff's theory deals with behavioral and linguistic aspects of polite communication and is therefore also suitable for tasks that do not require talking.

Politeness in the field of HRI has been explored in few studies. Research was performed with humanoid robots (receptionist robot (Salem et al. 2013), healthcare service (Lee et al. 2017), adaptive expressive robot (Ritschel et al.)) and in animation (gatekeeping robot (Inbar and Meyer 2019)). Most of those studies were based on Brown and Levinson's politeness theory (Brown et al. 1987). In the receptionist study, HRI performance was not influenced by the no-polite and polite behavior (Salem et al. 2013). The adaptive expressive robot had mixed responses from the participants based on gender. While males had a good perception of polite behavior, females were of quiet opposite view. However, regardless of demographics, all participants in the gatekeeper task preferred the polite behaving robot (Inbar and Meyer 2019). With Nao used as a healthcare service provider, it was found that the polite gesture with direct command (verbal no-polite behavior) was preferred over no gesture and indirect commands (verbal polite) (Lee et al. 2017). Another study investigated the influence of impolite behavior in an exercise task with a humanoid robot (Rea and Schneider 2021). Impolite behavior was defined as the "face threatening behavior" i.e. making a negative comment which could impact the social image of the person. This impolite behavior led to an increase in the overall performance of the user since it was compelling and competitive.

In the current research, we investigated the effect of politeness in a task that involved collaboration between a human and a manipulator robot. Based on previous research (Kumar et al. 2019), we formulated only two different levels of politeness –no-polite and polite behaviors. In the no-polite condition, Lakoff rules are not implemented. We use the term no-polite to contrast it with the other condition, yet the robot behavior in this condition is closer to being "strict" rather than the "impolite" version used in [16]. In the polite behavior condition, all three sub-rules were employed.

**Post-trial Questionnaire**
To what extent do you agree with the following statements regarding the interaction with the robot in this scenario? (1-5)
1 represents "strongly disagree" and 5 represents "strongly agree"
1. The interaction with the robot was enjoyable
2. The interaction of this robot could assist disabled people.
3. I felt the robot could be trusted
4. I think that people who will be assisted by the robot will enjoy this interaction
5. For this question please select number 2
6. People can trust the robot
7. I am satisfied with the way the robot communicated with me

**Final Questionnaire**
1. Did you feel a difference between the scenarios? Yes / No
   If so, what was the difference?
2. Which scenario is best for bringing the colored cubes? Right / left / right middle / left middle
3. Which scenario is the least desirable for bringing the colored cubes? Right / left / right middle / left middle
4. In what scenario would you say that the robot was the most polite? Right / left / right middle / left middle
5. In what scenario would you say that the robot was the least polite? Right / left / right middle / left middle

Figure 1. Post-trial and final questionnaire

This study also addresses the effect of mistakes made by the robot on the overall HRI performance comparing polite and no-polite robots. In the imperfect, realistic world of human-human interaction, errors done by "superior person with intellect" make the interaction likeable (Aronson et al. 1966). As social robots are penetrating our society, research must address also erroneous interactions between a human and a robot as suggested in (Mirnig et al. 2017). Despite the common perception that robots perform actions error-free (Bruckenberger et al. 2013), reality has shown that robots may err. Honig and Gilad (Honig and Oron-Gilad 2018) in their detailed survey proposed a taxonomy for various kinds of failures in HRI. They listed a large number of errors, which can happen during the interaction. Since our task is collaborative (detailed in the next section), we focused on an error caused by wrong actions committed by the robot.

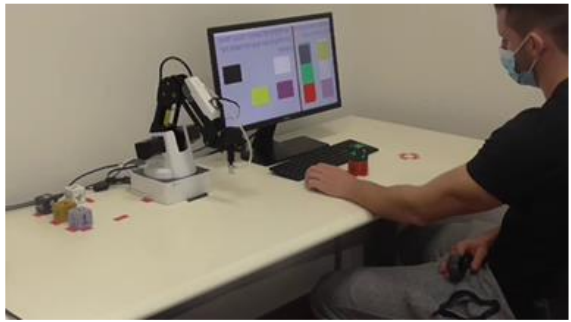

Figure 2. Participant interacting with the robot

There have been numerous studies evaluating the effect of erroneous robot behavior on the user

(Salem et al. 2015)(Ragni et al. 2016)(Lee et al. 2010)(Short et al. 2010). The reliability and trustworthiness of a faulty robot were less as compared to a faultless robot (e.g., (Salem et al. 2015), (Ragni et al. 2016)). However, objective task performance (e.g., a person following the robot) was not affected. An erroneous robot was perceived as less competent, less intelligent, inferior, and less reliable than an error-free robot (Ragni et al. 2016). In another study with a non-task related error, the faulty robot was more likable by the participants and the users did not perceive the robot as less intelligent (Mirnig et al. 2017). However, a service robot that made an error had a strong negative impact on people's ratings of the service quality (Lee et al. 2010). A robot playing a game (rock, paper, and scissors) with participants and producing wrong gestures was perceived as cheating behavior while verbal cheating was perceived as malfunctions by the user (Short et al. 2010).

This research focuses on the importance of politeness as a maxim to be included in the domain of HRI. It is also noted that the world of social robotics will encounter various kinds of errors during the interaction. It would be worthy to explore how politeness maxims affect the collaborative task. It should be noted that research pointed out that polite behavior helps in better mitigation practices in case of error (Sim 2021). However, it would be interesting to explore how a polite robot is perceived when it makes an error. Will the user be more frustrated toward the polite robot when making a mistake as compared to the no-polite one?

Methodology

A human-robot collaborative cube arranging task was developed. The user was asked to arrange six different colored cubes according to instructions provided through a graphical user interface (GUI). A four degrees of freedom DOBOT magician provided cubes to a user who did the actual arrangements. The robot and GUI were programmed in Python with two different politeness levels (no-polite and polite) and two different behaviors (correct, mistake) resulting in four different scenarios:

- Polite - Correct: The robot would first greet the user (adhering to the "*Be friendly*" politeness sub-rule). Then provide the user with options of different colored buttons corresponding to the different colored cubes (adhering to the "*Give option*", "*Don't impose*" rules).
- No-polite - Correct: The robot would bring the colored cubes in an order that would be easy for the user to arrange, but would not exhibit any of the polite behavior rules followed by the Polite robot.
- Polite - Mistake: The robot would first greet the user ("*Be friendly*"). Then the user would be provided with options of different colored buttons corresponding to the different colored cubes ("*Give options*", "*Don't impose*"). However, in this condition, the robot would make mistakes while bringing the cubes.
- No-polite - Mistake: The robot would bring the colored cubes, but not in the required order for the arrangement.

A. Hypotheses

We expect the following effects of polite behaviors and error actions of the robot on user's perceptions:

H1: Polite behavior will improve reactions to interactions with the robot on all measures (enjoyment, satisfaction and trust) relative to no-polite behavior.

H2: Error actions of the robot will have a negative impact on reactions to interactions with the robot on all measures compared to an error-free robot.

B. Procedure

Nineteen engineering students (average age $25.79 \pm 1.03$; 13 females; 6 males) were recruited through an ad on the department's website. Participants were compensated by 1 credit score in one of their courses. The experiment was approved by the university's ethical committee.

When the participants arrived at the experimental area they were asked to relax and were handed over the consent form. After taking their consent they were given a preliminary background questionnaire regarding usage of technology in daily life. After filling the questionnaire, they were briefed about the experiment. Each participant performed all four scenarios. After each scenario, they were asked to fill a post-trial questionnaire as shown in Fig.1 with statements regarding their current interaction with the robot related to enjoyment, satisfaction and trust. The answers to the questionnaire's items were scaled from 1 to 5, in which 1 represents "strongly disagree" and 5 represents "strongly agree". At the end of the experiment, there was a short final questionnaire as shown in Fig. 1 comparing all four scenarios. The whole experiment was recorded through a Panasonic video camera and took on average 30 minutes.

C. Experimental design

Each participant interacted with the robot via a GUI (Fig. 1). The four different scenarios were represented by four different buttons named "Left", "Left middle", "Right Middle" and "Right" which were not revealed to the user (no-polite-mistake, no-polite-correct, polite-mistake, polite-correct). The participants were briefed that the naming was just for the different scenarios and that they could choose any one of them in any order. The participants selected the order of execution which depended on the button they selected (this ensured a non-controlled, mixed order of execution).

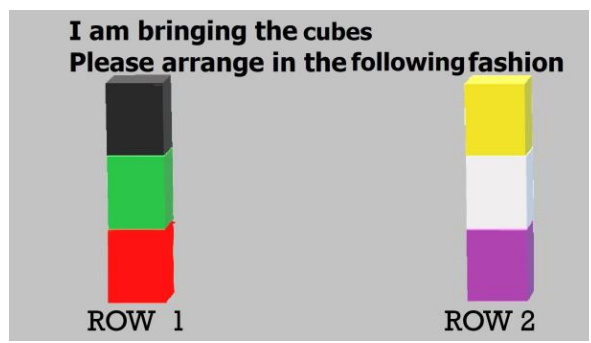

Figure 3. Instruction on GUI for arranging the cubes

The participants had to arrange the colored cubes according to the instructions displayed to them (example shown in Fig. 3). In the polite conditions, the buttons for the colored cubes are shown in Fig. 4. After pressing the colored button, the instruction page shown in Fig. 3 pops up and in parallel, the robot brings a colored cube to a preset position near to the participant as shown in Fig. 2. In the no-polite level after pressing the level button ("Left" or "Left middle") the instruction page shown in Fig. 3 pops up and the robot brings the colored cube according to the aforementioned scenarios. The process continues until all cubes were brought by the robot.

D.  Analysis

The dependent measures were the user's perceptions of the robot (enjoyment, satisfaction and trust) that were assessed through the post-trial questionnaire. Each measure was represented by two items in the post-trial questionnaire (enjoyment: questions 1+4, satisfaction: questions 2+7, trust: questions 3+6). The questionnaire included an additional question that was included as an instructional manipulation check to ensure the participants were filling the questions honestly. In this item, the participants had to fill a particular number in response to a statement.

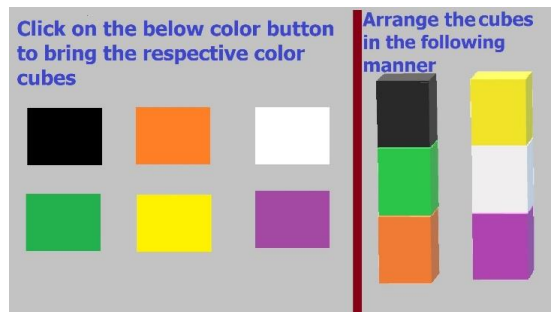

Figure 4. Different colored buttons required to press for different colored cubes

For initial analysis the mean and the standard deviation of all the responses for each of the four independent variables (polite - correct, polite - mistake, no-polite - correct, no-polite - mistake) was evaluated.

A Kolmogorov-Smirnov test indicated that the data was not normally distributed, hence we used non-parametric tests. Friedman test was performed to compare between the independent variables. The level of significance was adjusted by Bonferroni correction to 0.017. On obtaining a significant result, a post hoc test was conducted by Wilcoxon signed-rank test to determine the significance within the subgroups of the independent variables.

For each dependent measure, the mean of the responses from the participants in the polite-correct and polite-mistake was evaluated and termed as 'polite' behavior. Similarly, this was done for no-polite-correct and no-polite-mistake termed as a 'no-polite' behavior. The same process was repeated to obtain 'correct' (taking mean of responses from polite-correct and no-polite correct) and 'mistake' behaviors (taking mean of responses from polite-mistake and no-polite-mistake). This data was then analyzed through the Wilcoxon rank test within the group.

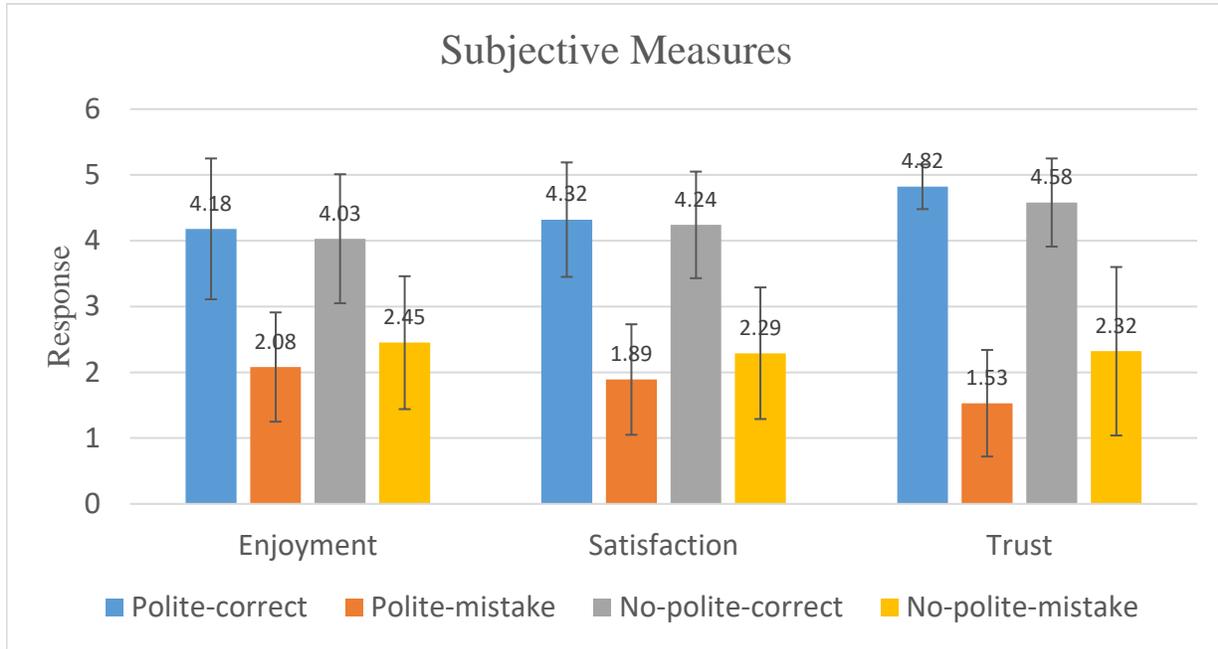

Figure 5. Mean and standard deviation of the dependent variables

All analyses were conducted in R studio software.

Results

All participants filled the instructional manipulation tests correctly, i.e., selected "2" on the rating scale.

The means and the standard deviations of the measures are described in Fig. 5. The Friedman test revealed significant differences for all the dependent variables i.e., enjoyment ($\chi2$ (3) = 38.37, p <0.001), satisfaction ($\chi2$ (3) = 48.60, p <0.001) and trust ($\chi2$ (3) = 46.46, p <0.001). Results of the post hoc analysis conducted to further investigate the differences among the conditions are reported in Table III (for clarity, only z-value and p-value are presented).

TABLE I. MEAN AND STANDARD DEVIATION OF HRI MEASURES OF MISTAKE AND CORRECT BEHAVIOR

| Subjective measures | Mistake | Correct |
| --- | --- | --- |
| Enjoyment | 2.26 ± 0.72 | 4.11 ± 0.78 |
| Satisfaction | 2.09 ± 0.69 | 4.28 ± 0.68 |
| Trust | 1.92 ± 0.69 | 4.70 ± 0.43 |

The mean and standard deviation of the correct and mistake behaviors (each for both polite and no-polite) of the robot are illustrated in Table I. Wilcox signed rank test revealed that there were significant differences in enjoyment (Z=0.81, p < 0.001), satisfaction (Z=0.81, p <0.001) and trust (Z = 0.81, p <0.001).

The mean and standard deviation of the no-polite and polite behaviors (each for both correct and mistake) are demonstrated in Table II. The test revealed that there were no significant differences in enjoyment (Z=0.02, p = 0.53), satisfaction (Z=0.03, p = 0.45) and trust (Z=0.21, p = 0.17).

TABLE II. MEANS AND STANDARD DEVIATIONS OF THE HRI MEASURES OF NO-POLITE AND POLITE BEHAVIORS

| Subjective measures | No-polite | Polite |
|---|---|---|
| Enjoyment | 3.24 ± 0.85 | 3.13 ± 0.75 |
| Satisfaction | 3.26 ± 0.77 | 3.11 ± 0.69 |
| Trust | 3.45 ± 0.81 | 3.17 ± 0.43 |

The analysis of the final questionnaire showed that all participants felt the difference between the four scenarios. 52.63 % of the participants preferred the polite - correct behavior of the robot while 36.84% of the participants preferred the no-polite - correct version and the rest preferred the polite - incorrect behavior. 68.42% of the participants least preferred the polite - mistake behavior, 26.31% of the participants least preferred the no-polite - mistake behavior and the rest least preferred the no-polite – correct behavior. 89.74% of the participants perceived the polite - correct behavior as the politest version of the robot while the rest perceived the no-polite - correct and the polite - mistake equally. 57.89% of the participants ranked the no-polite - correct behavior as the least polite version of the robot and 42.11% of the participants mentioned that the polite - mistake was the least polite behavior.

TABLE III. POST HOC ANALYSIS

| Different groups | Enjoyment | Satisfaction | Trust |
|---|---|---|---|
| No-polite - correct – No-polite - mistake | Z=0.80, p < 0.001 | Z=0.77, p < 0.001 | Z=0.77, p < 0.001 |
| No-Polite - correct – Polite - correct | Z=0.03, p = 0.55 | Z=0.12, p = 0.70 | Z=0.26, p = 0.12 |
| No-polite - mistake- Polite - correct | Z=0.75, p < 0.001 | Z=0.81, p < 0.001 | Z=0.79, p < 0.001 |
| No-polite correct – | | | |

Discussion

The effect of polite behaviors of the robot did not have any impact on the users' perception of the robot (H1 is rejected). This finding is in contradiction to the previous results (Kumar et al. 2019) in which the users (only engineering students which are in the similar age group) preferred the polite robot. This might have been due to the difference in the task – in previous research the participants were passive; in the current work the participants actively worked in collaboration with the robot to fulfill the task; they might have been too concentrated on this collaboration and less focused on the interaction.

The participants were more concerned about the robot making errors (H2 is supported). It could be noted that contrary to (Mirnig et al. 2017), the user's perception about the robot making an error was negative.

It could be argued that the participants have a mental model according to which a robot needs to do its job perfectly and therefore they were less concerned about the politeness of the robot and were more focused on its performance. Nearly half of the participants responded that the least preferable mode of interaction was the polite - incorrect behavior. It could be argued that people get more frustrated with a polite robot if it makes mistakes, or that robot's errors are considered impolite. It must be noted that in this experimental design, the novelty effect could have played an important role. Since the robot is perceived as being a perfect machine, this could affect the results. Hence, participants should be made aware that social robots can make mistakes.

Conclusion and future work

In this user study, polite maxims of human-human interaction were employed in a human-robot collaborative task. Based on Lakoff's theory, two different levels of polite behavior were designed. To introduce a more realistic behavior, errors were introduced into the robot's actions in both levels, creating four different experimental scenarios. Results revealed that the polite behavior did not impact the HRI performance. However, it was also observed that the mistakes done by the robot during the interaction had a negative impact on users' perceptions of the robot.

It could be argued that the reason for the impact on the dependent measures is having a mental model of a perfect-functioning robot. Future studies should address the novelty effect of an erring robot on participants, by adjusting their mental model of robots to be accustomed to a robot making errors.

## Acknowledgments

This research was supported by the EU funded Innovative Training Network (ITN) in the Marie Skłodowska-Curie People Programme (Horizon2020): SOCRATES (Social Cognitive Robotics in a European Society training research network), grant agreement number 721619. Partial support was provided by Ben-Gurion University of the Negev Agricultural, Biological and Cognitive Robotics Initiative, and the Rabbi W. Gunther Plaut Chair in Manufacturing Engineering.